\documentclass[11pt]{article}
\usepackage{rldmsubmit,palatino}

\usepackage{MT_packages}
\usepackage{MT_commands}

\title{Memory Allocation in Resource-Constrained Reinforcement Learning}

\author{
Massimiliano Tamborski
\\
School of Informatics\\
University of Edinburgh \\
\texttt{m.tamborski@sms.ed.ac.uk} \\
\And
David Abel\\
School of Informatics \\
University of Edinburgh \\
\texttt{david.abel@ed.ac.uk} \\
}

\begin{document}

\maketitle

\begin{abstract}

Resource constraints can fundamentally change both learning and decision-making. 
We explore how memory constraints influence an agent's performance when navigating unknown environments using standard reinforcement learning algorithms. 
Specifically, memory-constrained agents face a dilemma: how much of their limited memory should be allocated to each of the agent's internal processes, such as estimating a world model, as opposed to forming a plan using that model?
We study this dilemma in MCTS- and DQN-based algorithms and examine how different allocations of memory impact performance in episodic and continual learning settings. 
\end{abstract}


\startmain 

\section*{Introduction}

Humans have the ability to adapt to unknown situations. We can go to Dublin for RLDM and find our hotel, a restaurant, and the conference venue, even if we have never been there. Our behaviour may be sub-optimal---we may not take the shortest path---but we complete our goals within a reasonable time, despite limitations on our available resources such as memory or thinking time.
What cognitive tools do \textit{resource-constrained} learning algorithms need to realise this flexibility?

While Reinforcement Learning (RL) has shown great promise in training agents that autonomously interact with their surroundings \citep{kaufmann2023champion_level_drone_racing_using_DRL,wu2022daydreamer}, achieving autonomy is sometimes perceived as merely a matter of scaling up the agent's computational resources \citep{sutton2019bitter_lesson} rather than addressing it through fundamental research at the intersection of informatics and cognitive science. The \textit{Big World Hypothesis} \citep{dong2022simple-agent-comlex-env, javed2024the_big_world_hypothesis_and_its_ramifications_for_ai} suggests that in some especially interesting cases, the world is too complex for any agent to model perfectly. It is therefore important to understand RL problems involving \textit{simple agents} subject to resource constraints, inspired by the perspectives of bounded rationality~\citep{herbert1955a_behavioural_model_of_rational_choice, abel2019concepts_in_bounded_rationality_perspectives_from_rl}. 
In particular, we here explore the following question: 

\begin{quote}
    \begin{center}
        \textit{How do reinforcement learning algorithms effectively contend with memory constraints?}
    \end{center}
\end{quote}

\begin{figure}[htbp]
    \begin{minipage}{0.54\textwidth}
        \centering
        \includegraphics[width=\textwidth]{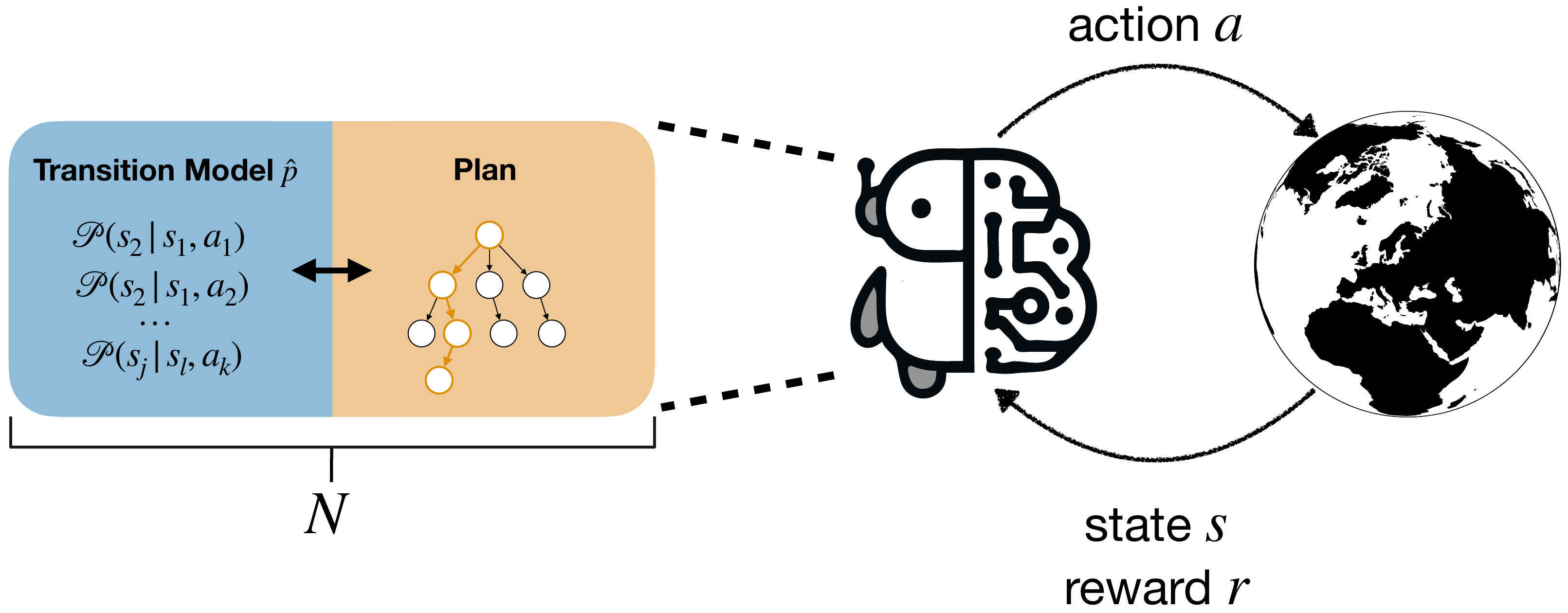}
    \end{minipage}
    \begin{minipage}{0.45\textwidth}
        \caption{Agent-environment loop for an agent who needs a \textit{transition model} $\hat{p}$ and a \textit{plan} to act (e.g., an MCTS agent). The rectangle containing the blue and orange areas represents the agent's total memory. There is a trade-off between having a more accurate estimate of the transition model or a more detailed plan. The double arrow between the blue and orange areas indicates that we can choose agent designs that commit to a different trade-off of these quantities.}
        \label{fig:methodology_feedback_loop_with_bounded_memory}
    \end{minipage}
\end{figure}

\paragraph{Problem: Memory-Constrained RL.} To address this question, we highlight a puzzle that any memory-constrained agent implicitly confronts. We consider the case in which a memory-constrained agent interacts with a finite Markov decision process (MDP; \citeauthor{puterman2014mdp}, \citeyear{puterman2014mdp}).
We denote an MDP as a tuple, $(\mathcal{S}, \mathcal{A}, r, p, \gamma)$, with $\hat{p}$ an estimate of the transition model and $\pi : \mathcal{S} \rightarrow \mathcal{A}$ a policy.
Our central assumption is that the agent only has $N \in \mathbb{N}$ total units of memory available to operate. Consider a model-based agent: it uses (1) an estimate of the transition model, $\hat{p}$, and (2) an internal plan over that transition model. The agent must then allocate a portion of its $N$ units of memory to the model estimate, $N_{\hat{p}}$, and some to the plan, $N_{\pi}$, where $N_{\hat{p}} + N_{\pi} = N$.
The agent then faces a dilemma: how should its limited memory be allocated between its model ($N_{\hat{p}}$) as opposed to its plan ($N_{\pi}$) so as to maximise performance?
If $N_{\hat{p}} \approx N$, the agent is afforded a more accurate world model but fewer units for representing its plan. If $N_{\pi} \approx N$, then the agent has the \textit{memory} to compute a longer plan but cannot \textit{actually} simulate what happens since the world model is limited. \Cref{fig:methodology_feedback_loop_with_bounded_memory} depicts this memory allocation problem, which can easily be extended to POMDPs \citep{kaelbling1998pomdp}, as in Question Two below.

For example, in Monte Carlo Tree Search (MCTS; \citeauthor{browne2012survey}, \citeyear{browne2012survey}), each node in the tree is a state $s$ and uses one unit of $N_{\pi}$, while one transition $\tau=(s_t,a_t,r_t,s_{t+1})$ one unit of $N_{\hat{p}}$. In MCTS, $\hat{p}$ is used during the expansion and simulation phases to determine how the state changes after taking an action. 
The formalism abstracts from the \textit{precise} definition of what constitutes a ``unit" to make the argument clearer and independent of any hardware or implementation detail.

\section*{Experiments} 

We first explore the trade-off the MCTS agent above faces. Second, we demonstrate how memory allocation can be generalised to different problems. Specifically, we ask the following two questions:
\begin{enumerate}
    \item \textit{How much does the quality of data given to an MCTS resource-constrained agent affect its performance?}
    \item \textit{Is the memory-allocation problem relevant to quantities other than $N_\pi$ and $N_{\hat{p}}$?}
\end{enumerate}

\paragraph*{Question One: Resource-Constrained MCTS.} To address the first question, we assume the agent has access to a dataset of transitions, $\mc{D} = \{\tau_1, \tau_2, \dots, \tau_n\}$ collected by an unknown policy, where $n$ may be larger than $N_{\hat{p}}$. In a streaming fashion, the agent can see all of these transitions once and then must select (here, randomly) $N_{\hat{p}}$ of them. We use MCTS to test how a simple and efficient algorithm performs given different memory capacities---that is, different settings of $N_{\hat{p}}$ and $N_{\pi}$---and dataset qualities. We set the memory budget to $N=500$ and we vary $N_{\pi}$ between $[0, N]$, with $N_{\hat{p}}=N-N_{\pi}$. The agent then builds $\hat{p}$ using a maximum-likelihood estimator (MLE) and $\pi$ using $\hat{p}$ and MCTS. We then use this plan to evaluate the agent. \Cref{alg:resource_constrained_agent} summarises the process for MCTS, but can easily be generalised to other settings. Although the algorithm suggests that the \textit{agent} chooses the memory split to use (\Cref{algo1_lane1}), below, we test all combinations of $N_{\hat{p}}$ and $N_{\pi}$ to evaluate how different choices affect the agent's performance. While we set $N=500$, we anticipate that the results will remain similar as long as $N$ is small relative to the size of the environment.

\begin{algorithm}
\caption{Resource-constrained agent}
\label{alg:resource_constrained_agent}
\begin{algorithmic}[1]
\Require $\text{total memory } N, \text{transitions } D, \text{initial state } s_0$
\State $N_{\hat{p}}, N_{\pi} = memory\_split(N, D)$ \label{algo1_lane1} \LeftComment{1.55}{Divide the memory for the transition model $\hat{p}$ and the plan}
\State $\hat{p} = model(D, N_{\hat{p}})$ \LeftComment{4.95}{Use the dataset to create a transition model using MLE}
\State $\pi = f(\hat{p}, s_0, N_{\pi})$ \LeftComment{5.5}{Use e.g., MCTS and $\hat{p}$ to compute a policy}
\State \Return $\hat{p}, \pi$
\end{algorithmic}
\end{algorithm}

\begin{figure}[h]
    \centering
    \begin{minipage}[t]{0.20\textwidth}
        \centering
        \includegraphics[width=\linewidth]{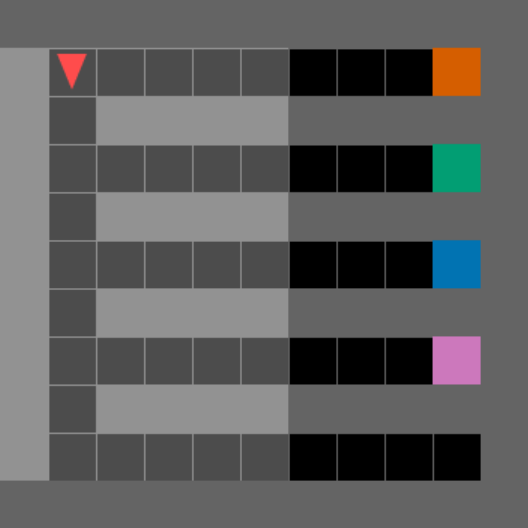}
    \end{minipage}
    \begin{minipage}[t]{0.38\textwidth}
        \centering
        \includegraphics[width=\linewidth]{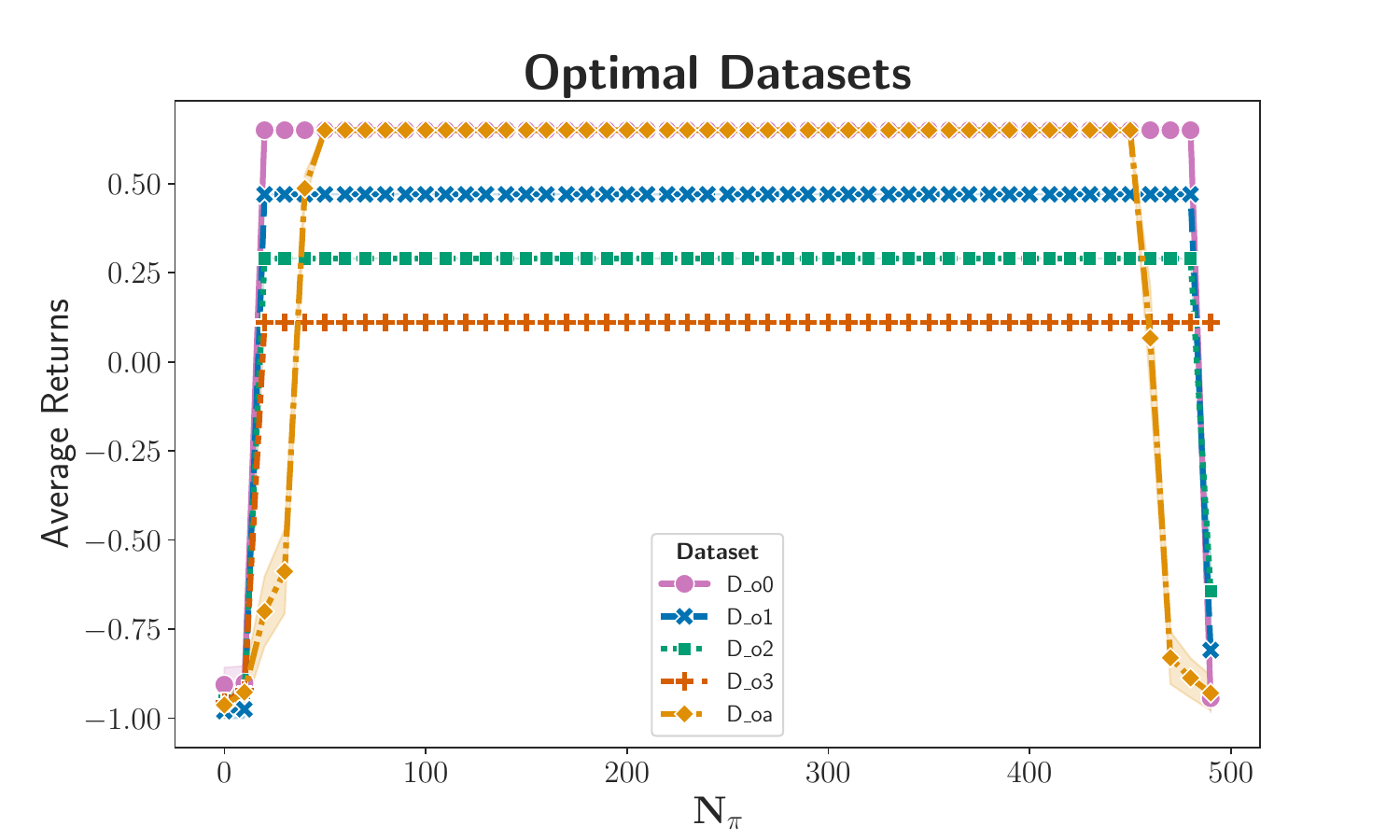}
    \end{minipage}
    \begin{minipage}[t]{0.38\textwidth}
        \centering
        \includegraphics[width=\linewidth]{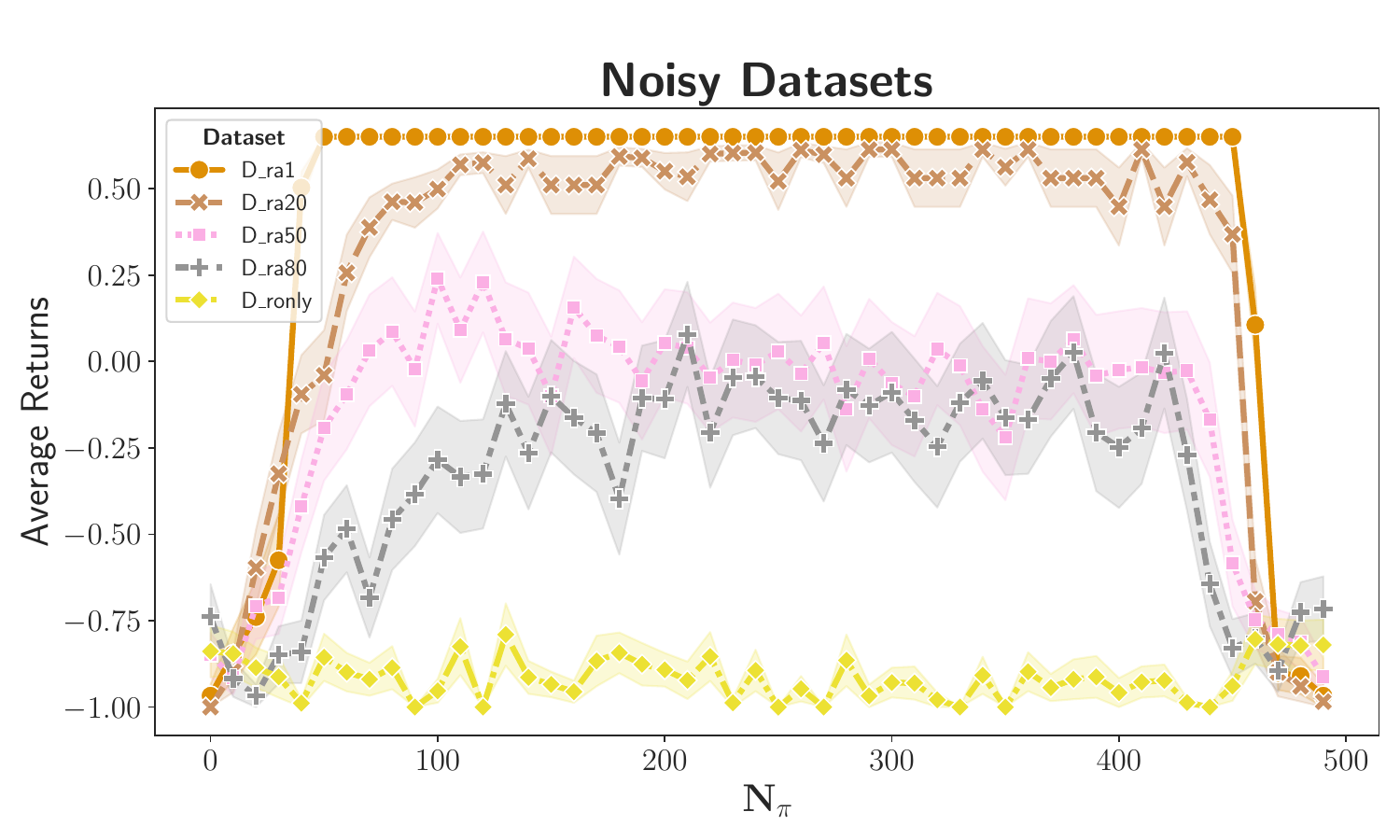} 
    \end{minipage}
    \caption{Evaluation return of policies trained on different datasets (lines; see the text for details) with varying memory sizes allocated to them (x-axis). \textit{Left:} CorridorEnv. \textit{Middle:} Policies trained on datasets only containing the trajectories to the goals. \textit{Right:} Policies trained on datasets containing the trajectories to all the goals and varying levels of noise.}
    \label{fig:mcts_results}
\end{figure}

\paragraph{Environment.} We use the MiniGrid \citep{MinigridMiniworld23} environment CorridorEnv proposed by \citet{arumugam2024satisficing_exploration_for_DRL}, illustrated in \Cref{fig:mcts_results}. The orange, green, blue, and pink squares indicate different goals the agent can reach. The reward for reaching each goal is proportional to its distance from the starting position (top-left corner). From orange to pink, the rewards are $0.2, 0.4, 0.6, \text{and } 0.8$. Each step carries a $-0.01$ penalty, and the episode ends when a goal is reached or after 100 steps. The highest achievable return is thus $0.65$, and the minimum is $-1$. The agent's memory capacity limits its ability to plan far ahead: the optimal goal to reach thus changes based on the capacity. 

\paragraph*{Results and Discussion.} \Cref{fig:mcts_results} shows the returns of policies trained on different datasets $\mathcal{D}$ and varying amounts of memory ($N_{\pi}$) allocated to them. Each point represents the return obtained after training a policy with \Cref{alg:resource_constrained_agent}, dataset $\mathcal{D}$ corresponding to the line the point belongs to, and $N_\pi$ corresponding to its x-coordinate.

Due to the stochasticity inherent to MCTS, we report the mean return averaged over 20 random seeds with confidence intervals indicating standard error. Dataset \texttt{$D_{o0}$} contains the trajectory to the best (pink) goal, \texttt{$D_{o1}$} to the second-best (blue) goal, and so on. Dataset \texttt{$D_{oa}$} contains the trajectories leading to each goal. Datasets \texttt{$D_{raX}$} contain the dataset \texttt{$D_{oa}$} and $X$ transitions collected by a random agent. \texttt{$D_{ronly}$} only contains transitions collected by this random agent.

As we vary the quality of data (e.g., $D_{o0}$ has higher quality than $D_{o3}$ as it leads to a higher reward), we observe that the returns tend to decrease, as expected. In both plots, when $N_{\pi}$ is low, the returns are close to zero. Only when $N_{\pi}$ is higher than the number of steps needed to reach a goal, can the agent reliably get positive rewards. When this condition is met, we see that the lines of the left plot flatten out before decreasing as $N_{\pi}$ approaches its maximum values. In this area, the agent has the \textit{memory capacity} to plan to the goal, but since $N=500$ and $N_{\hat{p}} = N-N_{\pi}$, the agent does not have enough memory to store the transitions needed to compute a plan. 

In the right plot, the curves do not completely flatten out. This behaviour is due to the noise present in the dataset. If $|\mathcal{D}| > N_{\pi}$, the $N_{\pi}$ transitions (randomly) sampled from $\mathcal{D}$ may not contain the trajectory to \textit{any} of the goals. Instead, the agent may waste resources by remembering useless transitions. In our setting, the agent cannot recover from data which does not contain at least one full (sub-)optimal trajectory, as the plan is only computed once at the beginning. This behaviour is demonstrated by the yellow line in the right plot, which always achieves the minimum returns possible. Since the agent cannot collect more data, it would need more resources than available to smooth the noise. 

An interesting trend is that all curves in the two plots (except for the yellow line) showed an inverse U shape. Performance roughly peaks when $N_{\pi} \approx 250 \approx N_{\hat{p}}$. The curves are not symmetric, hinting that the plan may need more resources than the model estimate. The observation makes sense: If $\mc{D}$ contains a lot of random transitions, the agent may need more planning capacity to account for ``dead branches" in the MCTS tree. The discount factor is $\gamma=1.0$. 

\paragraph*{Question Two: Beyond $N_\pi$ and $N_{\hat{p}}$.}
Working in a continual learning setting under partial observability,  \citet{anand2023prediction_control_in_CRL} propose PT-DQN, which splits the Q-Network in DQN \citep{mnih2015human_level_control_through_deep_rl_atari_deep_reinforcement_learning_dqn} into two, $Q = Q^{(P)} + Q^{(T)}$. Here, $Q^{(P)}$, the \textit{permanent} value function, captures knowledge shared across tasks, while $Q^{(T)}$, the \textit{transient} value function, captures specifics of the \textit{current} task. PT-DQN presents an interesting question: Anand and Precup assumed that the two Q-networks split the agent's memory evenly. Is this 50--50 \textit{PT-split} optimal for a resource-constrained agent? We kept $N=500$ as above---since the agent's total memory is determined by its hardware, \textit{not} the algorithm---and varied the PT-split. After testing different architectures on a standard DQN, the network with layers $[128,256,64,4]$ and a buffer of $52$ was the best one for the $N=500$ budget (we ignored the, fixed, output layer in the computations).

\begin{figure}[ht!]
    \centering
    \begin{subfigure}[b]{0.28\textwidth}
        \centering
        \includegraphics[width=\linewidth]{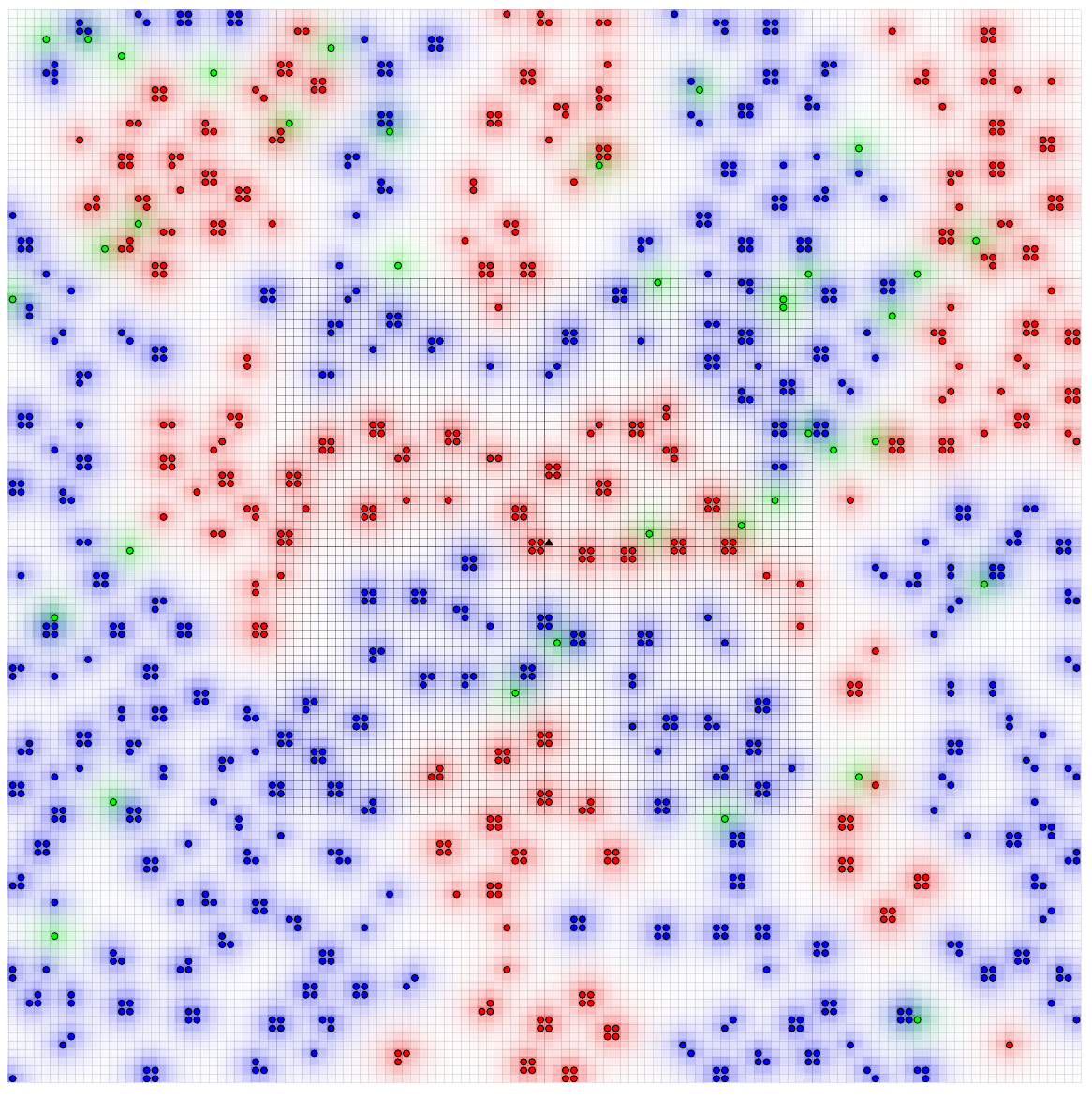} 
    \end{subfigure}
    \hfill
    \begin{subfigure}[b]{0.66\textwidth}
        \centering
        \includegraphics[width=\linewidth]{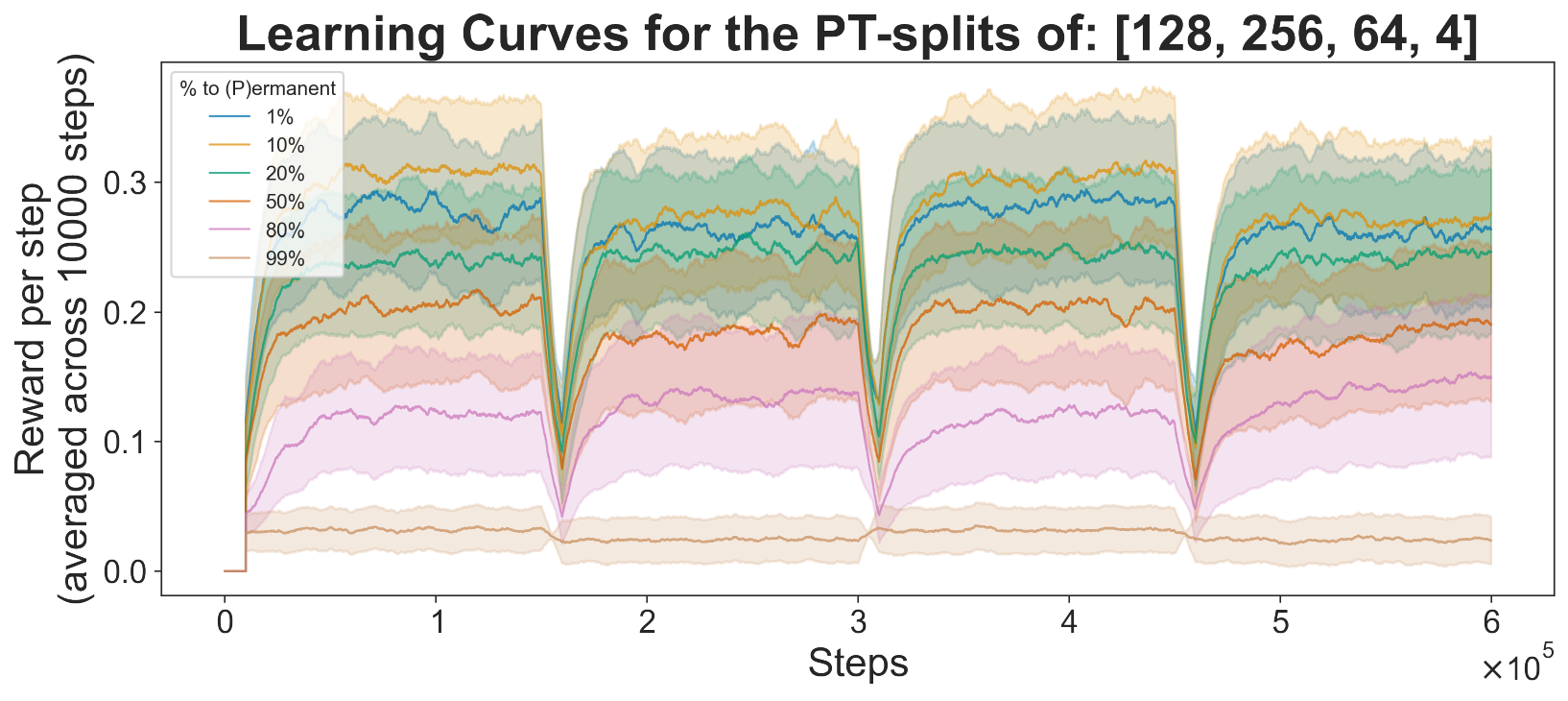}
    \end{subfigure}

    \caption{\textit{Left:} Jelly Bean World. \textit{Right:} Learning curves for different PT-splits in a PT-DQN agent with $N=500$. The number in the legend is the percentage of the total number of hidden units of the four-layer DQN $[128,256,64,4]$ allocated to the \textit{permanent} value function. The remaining neurons are for the \textit{transient value function}.}
    \label{fig:jbw_map_results}
\end{figure}

\paragraph*{Environment.} We used Jelly Bean World \citep{platanios2020jelly_bean_world_testbed_neverending_learning}, a 2D infinite grid illustrated in \Cref{fig:jbw_map_results}. Following Anand and Precup, green goals are uniformly distributed, each yielding a reward of $0.1$. The red and blue goals are clustered in groups and start with $-1$ and $2$ reward, respectively, but then swap these values every 150k steps. The agent receives an 11x11 RGB ego-centric image of the environment around itself. The action space is $\mc{A}=\{\text{up, down, right, left} \} $.

\paragraph*{Results and Discussion.} \Cref{fig:jbw_map_results} contains the per-step reward, averaged over 30 seeds, for different PT-splits of the $[128,256,64,4]$ network. Different PT-splits have a significant effect on the agent's ability to learn. Specifically, there seems to be a trend that a larger \textit{permanent} value function (increasing percentage in the legend) decreases the reward. The 50--50 PT-split proposed by Anand and Precup, for example, only reaches 0.2 per-step reward, while the best split (yellow line; 10\% of the hidden units for the permanent value function) reaches 0.3 reward. Another interesting result from the plot is that the yellow line is slightly higher than the blue one. While the confidence intervals are mostly overlapping, the results suggest that having a small \textit{permanent} value function might offer \textit{some} advantage over not having one.

\citet{tamborski2025bounded_agents_thesis} reports that when the same experiment is performed using the original network $[512, 256, 128, 4]$ and buffer size ($8k$) used by Anand and Precup, \textit{all} PT-splits result in a reward per step of around 0.3. While the memory-allocation problem may not be crucial to non-meaningfully constrained agents (as in the latter experiment, where $N >> 500$), it is crucial in real-world settings where the agent \textit{is} meaningfully constrained. 
For example, by allocating only 10\% of hidden units to the permanent value function (yellow line), the resource-constrained agent can almost equal the performance of the larger network. Another difference between the two networks is that the smaller one recovers faster than the larger one since the replay buffer is smaller, allowing it to replace outdated transitions more quickly.

\paragraph*{Related Work.} How resource-constrained agents can learn efficiently is an active area of research. \citet{abbas20selective_dyna_style_planning_under_limited_model_capacity} study how model-based agents can plan effectively despite memory constraints by selecting the parts of $\hat{p}$ they can trust. Some work proposes state abstractions ~\citep{li2006towards_unified_theory_of_state_abstraction_for_MDPs,konidaris2019on_the_nessecity_of_abstractions} as a way to simplify the high-dimensional RL decision-making problems, drawing inspiration from human learning ~\citep{ho2022people_construct_simplified_mental_representations_to_plan}. Since the plasticity vs stability dilemma is particularly evident in resource-constrained agents, \citet{javed2024the_big_world_hypothesis_and_its_ramifications_for_ai} and \citet{kumar2023continual_learning_as_computationally_contrained_rl,kumar2024the_need_for_a_big_world_simulator} suggest Continual Reinforcement Learning \citep{abel23definition-crl} may help resource-constrained agents learn.

\section*{Conclusion}

In this paper, we analysed the challenging task that resource-constrained agents face when allocating their limited memory. 
We first focused on MCTS in an episodic learning setting and found it struggles to produce useful policies when adapted to a memory-constrained setting. We then extended the analysis to neural networks and showed how memory allocation can allow resource-constrained agents to best use their resources and recover near-optimal performance.

Future directions include defining the nuances of a \textit{unit} and discovering how the agent can autonomously find the best memory allocation, possibly with a cognitive science lens. Given the importance of carefully allocating memory for both virtual and physical agents interacting in complex environments, we hope this simple analysis can spark discussion about some of the important questions and paths forward in memory-constrained RL. 

\bibliographystyle{abbrvnat}
\bibliography{refs.bib}

\begin{thebibliography}{23}
\providecommand{\natexlab}[1]{#1}
\providecommand{\url}[1]{\texttt{#1}}
\expandafter\ifx\csname urlstyle\endcsname\relax
  \providecommand{\doi}[1]{doi: #1}\else
  \providecommand{\doi}{doi: \begingroup \urlstyle{rm}\Url}\fi

\bibitem[Abbas et~al.(2020)Abbas, Sokota, Talvitie, and White]{abbas20selective_dyna_style_planning_under_limited_model_capacity}
Z.~Abbas, S.~Sokota, E.~Talvitie, and M.~White.
\newblock Selective dyna-style planning under limited model capacity.
\newblock In \emph{Proceedings of the International Conference on Machine Learning}, 2020.

\bibitem[Abel(2019)]{abel2019concepts_in_bounded_rationality_perspectives_from_rl}
D.~Abel.
\newblock Concepts in bounded rationality: perspectives from reinforcement learning.
\newblock \emph{Brown University Master thesis}, 2019.

\bibitem[Abel et~al.(2023)Abel, Barreto, Van~Roy, Precup, van Hasselt, and Singh]{abel23definition-crl}
D.~Abel, A.~Barreto, B.~Van~Roy, D.~Precup, H.~P. van Hasselt, and S.~Singh.
\newblock A definition of continual reinforcement learning.
\newblock In \emph{Advances in Neural Information Processing Systems}, 2023.

\bibitem[Anand and Precup(2023)]{anand2023prediction_control_in_CRL}
N.~Anand and D.~Precup.
\newblock Prediction and control in continual reinforcement learning.
\newblock In \emph{Advances in Neural Information Processing Systems}, 2023.

\bibitem[Arumugam et~al.(2024)Arumugam, Kumar, Gummadi, and Van~Roy]{arumugam2024satisficing_exploration_for_DRL}
D.~Arumugam, S.~Kumar, R.~Gummadi, and B.~Van~Roy.
\newblock Satisficing exploration for deep reinforcement learning.
\newblock \emph{arXiv preprint arXiv:2407.12185}, 2024.

\bibitem[Browne et~al.(2012)Browne, Powley, Whitehouse, Lucas, Cowling, Rohlfshagen, Tavener, Perez, Samothrakis, and Colton]{browne2012survey}
C.~B. Browne, E.~Powley, D.~Whitehouse, S.~M. Lucas, P.~I. Cowling, P.~Rohlfshagen, S.~Tavener, D.~Perez, S.~Samothrakis, and S.~Colton.
\newblock A survey of monte carlo tree search methods.
\newblock \emph{IEEE Transactions on Computational Intelligence and AI in games}, 2012.

\bibitem[Chevalier-Boisvert et~al.(2023)Chevalier-Boisvert, Dai, Towers, de~Lazcano, Willems, Lahlou, Pal, Castro, and Terry]{MinigridMiniworld23}
M.~Chevalier-Boisvert, B.~Dai, M.~Towers, R.~de~Lazcano, L.~Willems, S.~Lahlou, S.~Pal, P.~S. Castro, and J.~Terry.
\newblock Minigrid \& miniworld: Modular \& customizable reinforcement learning environments for goal-oriented tasks.
\newblock \emph{CoRR}, 2023.

\bibitem[Dong et~al.(2022)Dong, Van~Roy, and Zhou]{dong2022simple-agent-comlex-env}
S.~Dong, B.~Van~Roy, and Z.~Zhou.
\newblock Simple agent, complex environment: Efficient reinforcement learning with agent states.
\newblock \emph{Journal of Machine Learning Research}, 2022.

\bibitem[Ho et~al.(2022)Ho, Abel, Correa, Littman, Cohen, and Griffiths]{ho2022people_construct_simplified_mental_representations_to_plan}
M.~K. Ho, D.~Abel, C.~G. Correa, M.~L. Littman, J.~D. Cohen, and T.~L. Griffiths.
\newblock People construct simplified mental representations to plan.
\newblock \emph{Nature}, 2022.

\bibitem[Javed and Sutton(2024)]{javed2024the_big_world_hypothesis_and_its_ramifications_for_ai}
K.~Javed and R.~S. Sutton.
\newblock The big world hypothesis and its ramifications for artificial intelligence.
\newblock In \emph{Finding the Frame: An RLC Workshop for Examining Conceptual Frameworks}, 2024.

\bibitem[Kaelbling et~al.(1998)Kaelbling, Littman, and Cassandra]{kaelbling1998pomdp}
L.~P. Kaelbling, M.~L. Littman, and A.~R. Cassandra.
\newblock Planning and acting in partially observable stochastic domains.
\newblock \emph{Artificial Intelligence}, 1998.

\bibitem[Kaufmann et~al.(2023)Kaufmann, Bauersfeld, Loquercio, M{\"u}ller, Koltun, and Scaramuzza]{kaufmann2023champion_level_drone_racing_using_DRL}
E.~Kaufmann, L.~Bauersfeld, A.~Loquercio, M.~M{\"u}ller, V.~Koltun, and D.~Scaramuzza.
\newblock Champion-level drone racing using deep reinforcement learning.
\newblock \emph{Nature}, 2023.

\bibitem[Konidaris(2019)]{konidaris2019on_the_nessecity_of_abstractions}
G.~Konidaris.
\newblock On the necessity of abstraction.
\newblock \emph{Current Opinion in Behavioral Sciences}, 2019.

\bibitem[Kumar et~al.(2023)Kumar, Marklund, Rao, Zhu, Jeon, Liu, and Van~Roy]{kumar2023continual_learning_as_computationally_contrained_rl}
S.~Kumar, H.~Marklund, A.~Rao, Y.~Zhu, H.~J. Jeon, Y.~Liu, and B.~Van~Roy.
\newblock Continual learning as computationally constrained reinforcement learning.
\newblock \emph{arXiv preprint arXiv:2307.04345}, 2023.

\bibitem[Kumar et~al.(2024)Kumar, Jeon, Lewandowski, and Roy]{kumar2024the_need_for_a_big_world_simulator}
S.~Kumar, H.~J. Jeon, A.~Lewandowski, and B.~V. Roy.
\newblock The need for a big world simulator: A scientific challenge for continual learning.
\newblock In \emph{Finding the Frame: An RLC Workshop for Examining Conceptual Frameworks}, 2024.

\bibitem[Li et~al.(2006)Li, Walsh, and Littman]{li2006towards_unified_theory_of_state_abstraction_for_MDPs}
L.~Li, T.~J. Walsh, and M.~L. Littman.
\newblock Towards a unified theory of state abstraction for mdps.
\newblock \emph{AI\&M}, 2006.

\bibitem[Mnih et~al.(2015)Mnih, Kavukcuoglu, Silver, Rusu, Veness, Bellemare, Graves, Riedmiller, Fidjeland, Ostrovski, Petersen, Beattie, Sadik, Antonoglou, King, Kumaran, Wierstra, Legg, and Hassabis]{mnih2015human_level_control_through_deep_rl_atari_deep_reinforcement_learning_dqn}
V.~Mnih, K.~Kavukcuoglu, D.~Silver, A.~A. Rusu, J.~Veness, M.~G. Bellemare, A.~Graves, M.~Riedmiller, A.~K. Fidjeland, G.~Ostrovski, S.~Petersen, C.~Beattie, A.~Sadik, I.~Antonoglou, H.~King, D.~Kumaran, D.~Wierstra, S.~Legg, and D.~Hassabis.
\newblock Human-level control through deep reinforcement learning.
\newblock \emph{Nature}, 2015.

\bibitem[Platanios et~al.(2020)Platanios, Saparov, and Mitchell]{platanios2020jelly_bean_world_testbed_neverending_learning}
E.~A. Platanios, A.~Saparov, and T.~Mitchell.
\newblock {Jelly Bean World: A Testbed for Never-Ending Learning}.
\newblock In \emph{International Conference on Learning Representations (ICLR)}, 2020.

\bibitem[Puterman(2014)]{puterman2014mdp}
M.~L. Puterman.
\newblock \emph{Markov decision processes: discrete stochastic dynamic programming}.
\newblock John Wiley \& Sons, 2014.

\bibitem[Simon(1955)]{herbert1955a_behavioural_model_of_rational_choice}
H.~A. Simon.
\newblock A behavioral model of rational choice.
\newblock \emph{The Quarterly Journal of Economics}, 1955.

\bibitem[Sutton(2019)]{sutton2019bitter_lesson}
R.~S. Sutton.
\newblock The bitter lesson.
\newblock \emph{Incomplete Ideas (blog)}, 2019.
\newblock URL \url{http://www.incompleteideas.net/IncIdeas/BitterLesson.html}.

\bibitem[Tamborski(2025)]{tamborski2025bounded_agents_thesis}
M.~Tamborski.
\newblock Bounded agents in big worlds: A memory-constrained approach.
\newblock Master's thesis, The University of Edinburgh, 2025.

\bibitem[Wu et~al.(2022)Wu, Escontrela, Hafner, Goldberg, and Abbeel]{wu2022daydreamer}
P.~Wu, A.~Escontrela, D.~Hafner, K.~Goldberg, and P.~Abbeel.
\newblock Daydreamer: World models for physical robot learning.
\newblock \emph{Conference on Robot Learning}, 2022.

\end{thebibliography}
\end{document}